\documentclass[11pt,a4paper]{article}
\usepackage[hyperref]{eacl2021}

\usepackage{amsmath}
\usepackage{amssymb}
\usepackage{booktabs}
\usepackage{caption}
\usepackage{enumitem}
\usepackage{float}
\usepackage{framed}
\usepackage{graphicx}
\usepackage{latexsym}
\usepackage{subfigure}
\usepackage{times}
\usepackage{url}
\usepackage{xspace}
\graphicspath{ {figures/} }
\usepackage[ngerman]{babel}
\usepackage[utf8]{inputenc}
\usepackage[T1]{fontenc}
\usepackage{xcolor}

\usepackage{microtype}

\aclfinalcopy 

\newcommand{\mvs}{minimal variation set\xspace}
\newcommand{\mvss}{minimal variation sets\xspace}

\let\svthefootnote\thefootnote
\title{Neural language modeling of free word order\\ argument structure}

\author{Charlotte Rochereau \\
  Columbia University, USA \\
  {\tt\small cr3007@columbia.edu} \\\And
  Benoît Sagot \\
  Inria, France \\
  {\tt\small benoit.sagot@inria.fr} \\\And
  Emmanuel Dupoux \\
  ENS/CNRS/EHESS/Inria \\
  PSL Research University, France \\
  {\tt\small emmanuel.dupoux@gmail.com}}

\date{}

\begin{document}

\maketitle
\begin{abstract}
 Neural language models trained with a predictive or masked objective have proven successful at capturing short and long distance syntactic dependencies. Here, we focus on verb argument structure in German, which has the interesting property that verb arguments may appear in a relatively free order in subordinate clauses. Therefore, checking that the verb argument structure is correct cannot be done in a strictly sequential fashion, but rather requires to keep track of the arguments' cases irrespective of their orders. We introduce a new probing methodology based on \mvss and show that both Transformers and LSTM achieve a score substantially better than chance on this test. As humans, they also show graded judgments preferring canonical word orders and plausible case assignments. However, we also found unexpected discrepancies in the strength of these effects, the LSTMs having difficulties rejecting ungrammatical sentences containing frequent argument structure types (double nominatives), and the Transformers tending to overgeneralize, accepting some infrequent word orders or implausible sentences that humans barely accept.
\end{abstract}

\pretolerance 6000

\section{Introduction}

The development of natural language processing has produced neural models of remarkable capacity. Among them, long short-term memory (LSTM) networks \cite{hochreiter1997long} and Transformer architectures \cite{vaswani2017attention, devlin2018bert, raffel2020T5} trained in a self-supervised fashion provide excellent pretraining for a variety of downstream tasks. While considerable research is devoted in improving these architectures, it is also of interest to understand from a linguistic point of view how much of language knowledge these models really capture, and how do they compare to humans. 

Recent work has focused on exploring those representations and whether they correctly handle syntactic structures. \citet{linzen:2016}, \citet{gulordava:2018} and \citet{Lakretz2019TheEO} have studied number agreement, showing that LSTM LMs capture some hierarchical dependencies almost as accurately as humans. However, most of the work has focused on English, a comparatively poor morphosyntactic language. In morphologically richer languages, such as Basque, the verb number prediction task proves to be more challenging for linear models \cite{ravfogel2018can}.
We choose to focus on German, a morphosyntactically rich language with relatively free word order \cite{Uszkoreit1987Word}. Specifically, we probe neural LMs' syntactic capabilities on verb argument structures.

Verb argument structure provides languages with a way to link syntactic positions in a sentence (subject, direct object, etc) with semantic roles (agent, patient, etc). In many languages like English, verb argument structure is typically correlated to sentence position. But in other languages with freer word order, arguments can be shuffled (or \textit{scrambled}). Their function is then indicated by morphological markers. It is currently unclear whether neural LMs purely trained from surface statistics are able to capture this kind of structure, or whether additional information would be needed to provide some semantic grounding.

We setup a test of argument structure representation by presenting multiple pre-trained LMs with carefully constructed sets of sentences that either have the right set of arguments, or impossible sentences where one case is missing or duplicated. We use case order and semantic role permutations to control for unigram and positional statistics. If LMs are able to track argument structure irrespective of these permutations, they should assign higher acceptability scores to grammatical sentences than to ungrammatical ones.

Since at the level of the sentence, we study a global rather than local syntactic phenomenon, we depart from earlier work \cite{linzen:2016, gulordava:2018, marvin-linzen:2018, tran:2018} and do not compare pairs of sentences. Rather, we contrast a set of acceptable variations of the template sentence to a corresponding set of violations, which we call acceptable-unacceptable (A-U) \mvs. For each template sentence, we measure the model's ability to discriminate acceptable sentences from unacceptable ones using AUC-ROC curves, a well adapted metric for the classification of global variations. We then compare the model's performance to human evaluations, which set the gold standard for our dataset. Our results open up new methodological possibilities for systematically investigating large-scale grammatical variations and gradience in acceptability judgments.

We evaluate three LMs architectures on our dataset: LSTMs, Transformers (BERT and DistilBERT), and \textit{n}-gram baselines and compare their performances to human scores. We find that humans, LSTMs and Transformers all have a better than chance AUC-ROC score, with LSTM having a lower score than humans, and the Transformer higher scores than humans. Additional analyses of case orders and semantic role assignments reveal that human and model acceptability judgments are graded following implicit grammatical constraints, but the Transformer MLMs prioritize rules differently from humans and the LSTM. Furthermore, we find that the Transformers, unlike the LSTM and \textit{n}-gram baselines, do not rely on frequency cues for assigning sentence acceptability scores. Instead, they seem to generalize - possibly overgeneralize - syntactic rules to unseen structures in a non-human fashion. The human datasets and analysis scripts will be released upon manuscript acceptance.

\section{Background}

\subsection{Minimal pairs}
Acceptability judgments for recurrent networks have been investigated since \citet{allen1999:emergence}, who use closely matched pairs of sentences to investigate acceptable correctness. This approach has been adopted recently to assess the syntactic representations of LSTMs. \citet{linzen:2016} and \citet{gulordava:2018} use word probes in minimally different pairs of English sentences to study number agreement. To discriminate original (acceptable) sentences from nonce sentences, they retrieve the probabilities of the possible morphological forms of a target word, given the probability of the previous words in the sentence. In the sentence ``the boy \underline{is} sleeping'', the network has detected number agreement if \textit{$\textrm{P}(w = is) > \textrm{P}(w  = are)$}. This methodology has also been adapted by \citet{goldberg:2019} to non-recurrent models trained with a masked language-modeling objective (BERT and OpenAI GPT). Those works find that in the absence of many distractors or complex sentence features, recent Transformer LMs perform well at number-agreement in English. Using Boolean classifiers on minimal pairs, \cite{warstadt2019investigating} also test the syntactic knowledge of BERT MLM on negative polarity items (NPI) and show that BERT is better at detecting NPIs (``ever'') and NPI licensors (
``whether'') than their scope.

In contrast with approaches that seek to probe language models directly, other approaches involve fine-tuning representations to a specific syntactic task using a task-specific supervision signal. For instance, \citet{cola} introduce CoLA, a binary acceptability dataset whose example sentences are taken from linguistic publications. They train a classifier on top of frozen ELMo \cite{elmo} layers to assess performance at acceptability judgments. Later work \cite{devlin2018bert, cola_grammatical} has focused on fine-tuning an entire pre-trained model to the acceptability task, such as is done for BERT \cite{devlin2018bert}. Both of those paradigms do not directly evaluate syntactic ability but rather whether pre-trained representations can be effectively transferred to learn to solve specific syntax problems.

\subsection{Minimal variation sets}
In parallel to work focusing on word probe probabilities in minimal pairs, another closely related line of inquiry has investigated LMs' syntactic abilities using 2x2 interactions \cite{wilcox:2018,futrell:2019,wilcox2019structural}. Each sentence appears in 4 conditions, 2 acceptable and 2 less acceptable or unacceptable, reflecting the studied syntactic phenomenon and its violation. With this design, the authors examine the neural representations of several syntactic phenomena by measuring surprisal, the inverse log probability assigned by a model to a specific prediction.

We depart from these approaches since our test set encompasses whole sentence variations, such as argument reordering. Word probes are therefore less apt to capture such changes. Instead, we choose to follow \citet{marvin-linzen:2018} and \citet{tran:2018} in taking the more general approach of comparing whole sentence probabilities as our acceptability probe: the model should assign a higher probability to sentences that are acceptable to humans than to sentences that are unacceptable. 

Further, we extend the concepts of minimal pairs and 2x2 interactions to acceptable-unacceptable (A-U) minimal variation sets: instead of assessing the difference between two sentences, we compare |A| acceptable sentences to |U| minimally different yet unacceptable sentences. This allows us to assess whether models can capture the different degrees of acceptability between many minimal variations of a single sentence in a highly controlled setting. 

\subsection{Metric}
\begin{figure}[ht]
\centering
\includegraphics[width=.8\linewidth]{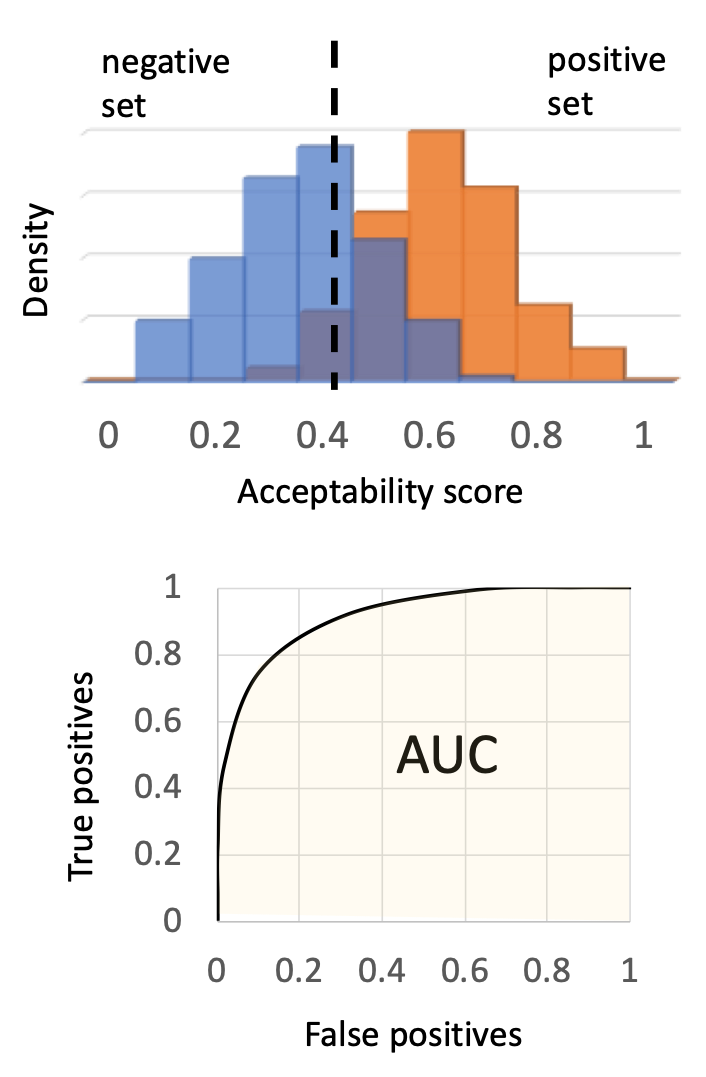}
\caption{(a) Acceptability scores for each sentence in a \textit{\mvs} provide a positive and a negative distribution. (b) ROC curve plotted for the separation of these two score distributions; the AUC is the area under the true-positive versus false-positive curve.
}
\label{fig:analogy_drawing}
\vspace{-1em}
\end{figure}

We use the AUC (Area Under the Curve) ROC (Receiver Operating Characteristics) as our classification metric (Figure \ref{fig:analogy_drawing}). The ROC is the plot of the true positive rate (y-axis) against the false positive rate (x-axis) for different probability thresholds. The AUC summarizes the model's ability to correctly predict the acceptability of input sentences. Higher the AUC, better the model is. An AUC of 1 (resp.~0) indicates that the model correctly labelled all (resp.~no) input sentences; 0.5 corresponds to chance level. We report the AUC for each (A-U) minimal variation set in our dataset for humans (our gold standard) and models.

\subsection{Verb Argument Structures}
Our investigation of verb argument structure representations in humans and neural models is most closely related to the work of \citet{Keller2001Gradience}, which explores the gradience of acceptability judgments. The author specifically studies verb argument orders in German subordinate clauses, in which NPs' order can vary. Human evaluations of sentences in which NP arguments are permuted reveal that argument orders have different acceptability degrees; these are influenced by the violation of grammatical constraints. Relevant constraints for our experiment are: i) $[+NOM]\prec[-NOM]$ (\textrm{NOMALIGN}) and ii) $[+DAT]\prec[+ACC]$ (\textrm{DATALIGN}), with $\prec$ denoting linear precedence. \textrm{NOMALIGN} means that clauses in which a nominative NP precedes other NPs will receive a higher acceptability degree. Its violation, ie a nominative NP comes after other NPs, produces less acceptable sentences. Similarly, \textrm{DATALIGN} means that clauses in which a dative NP precedes an accusative NP are more acceptable. 

Further, the author finds that grammatical constraints are ranked, i.e some constraints, when violated, lead to a more significant reduction in acceptability than others. They are also cumulative: the simultaneous violation of several constraints induces a higher degree of unacceptability than a single violation. In our experiment, successful neural LMs should assign higher scores to sentences which humans find acceptable (sentences that follow grammatical constraints) than to sentences which humans find unacceptable (sentences which violate grammatical constraints). This would be reflected by a high AUC (>> 0.5) for the corresponding (A-U) \mvss.

\section{Verb Argument Structure Dataset}
\label{sec:dataset}

\subsection{Templates} 

\begin{figure}[t]
\scriptsize
\begin{framed}    
\begin{flushleft}

  \textbf{\underline{Template:}}\;\;\; \\
 $[$NDA; ag1,pa2,re3$]$ \textit{Er wollte uns sagen, dass der$_N$ Soldat$_{\text{ag}:1}$ dem$_{D}$ Offizier$_{\text{pa}:2}$ einen$_{A}$ Brief$_{\text{re}:3}$ schreibt.}  
  (He wanted to tell us that the soldier writes a letter to the officer.)\\
  \textsc{50 templates} \\
\ \\

   \textbf{\underline{Acceptable permutations set:}}\\
  \ \\
\textbf{Permuting case order: (6 possibilities)}\\
$[$DNA; ag1,pa2,re3$]$ \textit{Er wollte uns sagen, dass \underline{dem Offizier} \underline{der Soldat} einen Brief  schreibt.} 
(He wanted to tell us that the soldier writes a letter to the officer.) \\
\ \\

  \textbf{Permuting role assignment:(6 possibilities)}\\
  $[$NDA; ag2,pa1,re3$]$ \textit{Er wollte uns sagen, dass \underline{der Offizier} \underline{dem Soldat} einen Brief schreibt.} (He wanted to tell us that the officer writes a letter to the soldier.) \\
\ \\

  \textbf{Permuting both}\\
 $[$DNA; ag2,pa1,re3$]$ \textit{Er wollte uns sagen, dass \underline{dem Soldat} \underline{der Offizier} einen Brief schreibt.} (He wanted to tell us that the officer writes a letter to the soldier.) \\~\\
  \textsc{$6 \times 6 = 36$ acceptable permutations per template} \\~\\

\textbf{\underline{Case violations sets:)}}\\
  \ \\
\textbf{Nominative case violation: (36 possibilities)}\\
 $[$NNA; ag1,ag2,pa3$]$ \textit{Er wollte uns sagen, dass \underline{der Soldat} \underline{der Offizier} einen Brief schreibt.} 
(He wanted to tell us that the soldier the officer writes a letter.) \\
\ \\

\textbf{Accusative case violation: (36 possibilities) }\\
 $[$NAA; ag1,pa2,pa3$]$ \textit{Er wollte uns sagen, dass der Soldat \underline{den Offizier} \underline{einen Brief} schreibt.} 
(He wanted to tell us that the soldier writes a letter the officer.) \\
\ \\

\textbf{Dative case violation: (36 possibilities)}\\
$[$NDD; ag1,re2, re3$]$  \textit{Er wollte uns sagen, dass der Soldat \underline{dem Offizier} \underline{einem Brief} schreibt.} 
(He wanted to tell us that the soldier writes to the officer to a letter.) \\
~\\
\textsc{$3 \times 36 = 108$ unacceptable permutations per template}
\end{flushleft}
\end{framed}

\caption{Construction of acceptable examples by permuting case assignments and  argument orders in template sentences. Some case orders may be more or less \textit{marked}; some role assignments more or less \textit{plausible}. For the construction of the unacceptable examples, we duplicate one of the cases, creating an impossible argument structure.}
\label{fig:examples}
    
\vspace{-1.5em}
\end{figure}

Our test sentences are generated from 50 German sentences (\textit{templates}). These templates all consist of a declarative main clause, where the verb is in second position, and a subordinate clause, which is verb final. The verb in the subordinate clause is ditransitive (such as \textit{vorstellen},``introduce'') and takes three NPs as arguments: a subject or \textit{agent} (ag), marked by the nominative case (N), a direct object or \textit{patient} (pa), marked by the accusative case (A) and an indirect object or \textit{recipient} (re), marked by the dative case (D). For simplicity purposes, we do not use any intervening element. In the template of Figure~\ref{fig:examples}, ``the soldier'' is the subject, ``a letter'' the direct object, and ``the officier'' the indirect object of the verb ``writes''.

We construct a dataset designed to expose impossible verb argument structures by manipulating the arguments' sequential case order (NAD, ADN...) and semantic roles (ag, pa, re) in subordinate clauses. Each lexical item is assigned a number (1, 2 or 3) to keep track of its position and semantic role. Items 1 and 2 are always human; item 3 is typically non-human (inanimate). This allows us to test whether models are able to capture syntactic dependencies when the positions and thematic relations of verb arguments vary. 

In German, NPs syntactic role is indicated by the morphological form of its constituents: determiners and nouns take different suffixes, if not completely different forms, according to their case assignment. However, feminine, neuter and all plural noun phrases (NPs) share common morphological forms. To avoid sentence duplicates within our dataset, we only use singular masculine NPs. 

\vspace{-0.5em}
\subsection{Acceptable Sets} 
To control for all possible argument orders and words syntactic roles, for each template, we permute (i)~the positions of the 3 verb arguments and (ii)~each NP's case assignment. There are $3$ verb arguments, leading to $6$ different position permutations. Similarly, they are 3 unique case assignments, leading to $6$ possible case assignments. By generating all such permutations, we create $6 \times 6 = 36$ acceptable sentences for each template. In Figure~\ref{fig:examples}, we show an example where only the positions of the subject and the indirect object are switched, which does not alter the meaning. We also show an example where only the case assignments of the subject and the indirect object are switched: ``the officer'' becomes the subject and ``the soldier'' the indirect object. We permute cases by retrieving the desired case markings from a dictionary mapping the vocabulary's nouns to their morphological forms. Case permutations change sentence meaning.

\subsection{Case Violation Sets} We construct unacceptable sentences using the same templates, by substituting one of the case assignments with another one already present in the sentence. This creates a grammatical violation: there are now 3 NPs and only 2 case assignments, one being duplicated. In Figure~\ref{fig:examples}, we show how we apply this to a template sentence to create acceptability violations. For each case violation, we generate 36 sentences containing a case violation from every template. Thus, from each of our 50 templates, we generate 36 acceptable variations and 108 unacceptable ones. Overall, our dataset comprises 7,200 sentences, of which 1,800 are labelled as acceptable and 5,400 as unacceptable.

\vspace{-0.6em}
\section{Methods}

\let\thefootnote\relax\footnote{Materials and code are available at \url{https://github.com/crochereau/probing-LM-syntax}.}
\addtocounter{footnote}{-1}\let\thefootnote\svthefootnote

\vspace{-1.5em}
\subsection{Human Evaluations}
\label{sec:human_eval}

\paragraph{Sentence Acceptability} To generate human acceptability evaluations for our dataset, we hire annotators proficient in German via Appen. We ask the annotators to assess sentence acceptability on a continuous scale from ``not natural'' to ``very natural''. This choice corresponds to a numeric value on a continuous scale from 0 (``not natural'') to 99 (``very natural'') which we use to compute human AUC values. During a warm-up phase, respondents are shown examples of sentences which are labelled as acceptable or unacceptable in our dataset. They are asked to judge how natural-sounding sentences are, irrespective of whether the situation described is likely or not.

Each respondent graded 216 sentences, with the following constraints: (i)~the respondent sees a maximum of 5 unique sentences from a single template, to avoid that past sentences impact future ratings; (ii)~25\% of the sentences shown are acceptable ones, mirroring the construction of the dataset; (iii)~sentences shown are randomly chosen among the 144 possibilities for each template, so that each user is exposed to a wide variety of case assignments, argument orders and acceptability violations; (iv)~no sentence is annotated twice. For comparison purposes, we normalize the ratings across participants. In total, we collect three annotations for each sentence of the dataset, which we average after normalization to obtain the sentence scores. We verify that the templates' average scores are not significantly different.

\paragraph{German Proficiency} To ensure that all annotators are proficient in German, we take the following steps: (i)~we only accept annotators from German-speaking countries; (ii)~annotators must pass a preliminary linguistic test; (iii)~instructions are given in German only; (iv)~filler sentences (sentences for which answers are known and obvious to proficient German speakers) are inserted throughout the annotation process to ensure annotators stay focused; (v)~we remove annotators whose average score for acceptable fillers is less than their average score for fillers that contain grammatical violations.

\paragraph{Individual Grading vs Pairwise Ranking} As noted, we do not ask humans to compare minimally different sentences, but rather to grade individual sentences. This setup differs from earlier work such as \citet{marvin-linzen:2018}, who show both sentences simultaneously and ask humans to pick the most acceptable one. This approach prevents humans from using the differences between the sentences to form an acceptability judgment; rather they must judge each sentence on its own. In doing so, the human setup is closer to that of LMs, which assign sentence scores without learning from differences between sentences.\\

\vspace{-1.5em}
\subsection{Language Models}

We probe language models with different architectures on our verb argument structure dataset and compare their performances to human scores. Sentence scores depend on models' architectures. 

\paragraph{LMs Scores} Traditional LMs, like \i{n}-grams or LSTM LMs, predict the next token given the previous context. The log probability of an input sentence is given by the chain rule:

\vspace{-1em}
\begin{equation*}
    \log \bf{P}_{LM}(w_{1}\dots w_{n}) = \sum_{i=1}^{n}{\log \bf{P}_{LM}(w_i}|w_{<i}).
\end{equation*}
\vspace{-1em}

\noindent By contrast, Transformers MLMs use both past and future tokens. Their bidirectional nature makes their use as LMs less obvious. Like \citet{Shin2019EffectiveSS} and \citet{salazar2020masked}, we use the \i{pseudo-log likelihood score} (PLL) proposed by \citet{Wang2019BERTHA}. We retrieve the words' conditional log probabilities by masking each sentence token one at a time. The model predicts the masked token using the other observed sentence tokens. The sentence score is obtained by summing the conditional log probabilities of the sentence masked words: 

\vspace{-1em}
\begin{equation*}
    \bf{PLL}(w_{1}\dots w_{n}) = \sum_{i=1}^{n}{\log \bf{P}_{MLM}(w_{\setminus{i}; \Theta}}). 
\end{equation*}
\vspace{-1em}

\noindent Each of these log probabilities can be read from the softmax outputs of the neural models, or directly estimated in the case of the unigram and bigram models. 

\paragraph{LSTM} We test the word-level German language model (German WordNLM) trained by \citet{hahn2019tabula}.\footnote{Models architecture (RNN: embedding size/layers/hidden size; Transformers: layers/hidden size/heads/parameters): LSTM: 100/2/1024, BERT: 12/768/12/110M, DistilBERT: 6/768/12/66M.} 
This model was trained on a 819M tokens from German Wikipedia. The vocabulary includes the 50K most frequent words in this corpus. The model reaches a perplexity of 37.96 on this dataset.

\paragraph{Transformers} We evaluate two pretrained Transformer models: German BERT (``bert-base-german-cased''), trained on 12GB of German data (including German Wikipedia)\footnote{\url{https://deepset.ai/german-bert}}, and German DistilBERT (``bert-base-german-dbmdz-cased''), trained on half the data used to pretrain BERT \cite{sanh2019distilbert}. Their vocabulary consists of 30K tokens.

\begin{table*}[t]
\centering\small
\begin{tabular}{lccccccc}
\toprule
Role assignment / & ag1, re2, & ag2, re1& ag1, re3, & ag2, re3, & ag3, re1, & ag3, re2, & \it Avg \\
Case order &  pa3 & pa3  & pa2 & pa1 & pa2 & pa1 & \it markedness\\
\midrule
NAD & 0.92 & 0.86 & 0.87 & 0.82 & 0.59 & 0.58 & \it 0.77 \\
NDA & 0.99 & 0.99 & 0.60 & 0.58 & 0.58 & 0.58 & \it 0.72 \\
DNA & 0.84 & 0.85 & 0.49 & 0.45 & 0.54 & 0.43 & \it 0.60 \\
AND & 0.68 & 0.75 & 0.65 & 0.64 & 0.44 & 0.48 & \it 0.61 \\
DAN & 0.65 & 0.62 & 0.40 & 0.47 & 0.56 &  0.59 & \it 0.55 \\
ADN & 0.65 & 0.65 & 0.45 & 0.40 & 0.57 & 0.58 & \it 0.55 \\
\it Avg plausibility & \it 0.79 & \it 0.79 & \it 0.58 & \it0.56 & \it 0.55 & \it 0.54 & \it 0.63 \\
\bottomrule
\end{tabular}
\caption{(1-6) \mvss human AUCs averaged by case order and ranked by increasing \textit{markedness} (NDA least marked) and decreasing \textit{plausibility} (ag1,re2,pa3 most plausible). AUCs > 0.5 mean that humans find sentences labelled as acceptable indeed more acceptable than sentences labelled as unacceptable.} 
\label{tab:humans_1to6}
\end{table*}

\paragraph{\textit{n}-grams} We consider two baselines: a unigram model and a bigram model with Laplace smoothing trained on the same German Wikipedia corpus. Our baselines cannot capture verb argument dependencies due to their limited window sizes.

\section{Results}
\label{sec:results}

\subsection{(1-6) \mvss}
We first report AUCs for 1-6 \mvss. For each sentence labelled as acceptable in our dataset there are 6 minimally different sentences labelled as unacceptable, each unacceptable sentence showcasing a doubled case. We compute the AUCs on these sets for human and model scores and average them by case order and semantic role assignments. Table \ref{tab:humans_1to6} shows the human results\footnote{Results for all models can be found in the Appendix.}. Case orders are ranked by increasing markedness, i.e how unusual case orders are. Semantic role assignments are ranked by decreasing plausibility. Human AUCs are almost 1 for nominative starting sentences and for agent and recipient as human entities. These results establish the canonical case orders (NAD, NDA) and semantic role assignments (ag: item 1, re: item 2 and vice versa).

Note that moderate human AUCs, like 0.65 for (ADN/Ag1,Re2,Pa3), indicate that such case order/semantic role permutations rarely (or never) occur in German, establishing that the sentence experimentally labelled as acceptable in our dataset is almost unacceptable to native speakers. The gradience in our human AUCs confirm the conclusions of \citet{Keller2001Gradience}: the possible case orders in German subordinate clauses have different acceptability degrees, reflecting linguistic constraints implicitly learned by German native speakers. We confirm the validity of the \textrm{NOMALIGN} rule: the earlier the nominative NP in the clause, the higher the AUC. The \textrm{DATALIGN} constraint is also verified for canonical role assignments (ag1,re2,pa3 and ag2,re1,pa3). We also confirm that \textrm{NOMALIGN} is stronger than \textrm{DATALIGN}: on average, case orders starting with nominative NP (NAD, NDA) reach the top AUCs, while those ending with a nominative NP (DAN, ADN), marking a violation of the \textrm{NOMALIGN} constraint, receive the lowest ones. In \ref{subsection:markedness}, we compare human and model AUCs in light of these linguistic constraints.

\subsection{Markedness}
\label{subsection:markedness}

Table \ref{tab:markedness} show AUCs on (1-6) \mvss averaged by decreasing markedness. As expected, \textit{n}-grams do not significantly differ from chance. Notable are the performances of the Transformers. Unlike the LSTM AUCs, which are on par with those of humans, BERT achieves much higher AUCs than humans for all case orders. Despite different acceptability degrees across case orders, BERT AUCs all near 0.8. DistilBERT's performances only slightly lag behind those of BERT. For both models, the best AUC is achieved for the most canonical case order (NAD). 

Interestingly, the grammatical constraints verified for humans are visible in the baselines and clearly account for the LSTM scores. The significant drop in performance of these models for case orders starting with an accusative NP shows that \textrm{DATALIGN} is even stronger for them than for humans. To the exception of \textrm{NOMALIGN} for DistilBERT, the Transformers do not seem to be influenced by these rules.

\begin{table}[t]
\centering\small
\begin{tabular}{l @{\hspace{0.5em}} c @{\hspace{0.5em}}  c @{\hspace{0.5em}} c@{\hspace{0.5em}} c @{\hspace{0.5em}} c @{\hspace{0.5em}} c}
\toprule
Case &        &       &       &       &       & Distil-\\
order & humans & unigr.&  bigr.&  LSTM &  BERT &  BERT \\
\midrule
NAD & 0.77 & 0.48 & 0.53 & 0.75 & 0.91 & 0.89 \\
NDA & 0.72 & 0.51 & 0.54 & 0.77 & 0.81 & 0.81 \\
DNA & 0.60 & 0.51 & 0.51 & 0.60 & 0.82 & 0.76 \\
AND & 0.61 & 0.48 & 0.45 & 0.39 & 0.80 & 0.73 \\
DAN & 0.55 & 0.52 & 0.61 & 0.56 & 0.78 & 0.60 \\
ADN & 0.55 & 0.52 & 0.39 & 0.35 & 0.81 & 0.70 \\
\it Avg &\it 0.63&\it 0.51&\it 0.51&\it 0.57&\it 0.82&\it 0.75\\
\bottomrule
\end{tabular}
\caption{(1-6) \mvss AUCs across humans and models averaged by case order and ranked by increasing \textit{markedness} (NDA least marked).}
\label{tab:markedness}
\end{table}

\subsection{Semantic Plausibility}
\label{subsection:plausibility}

In Table \ref{tab:plausibility}, we show the human and model AUCs on (1-6) \mvs averaged by decreasing semantic plausibility. This separate analysis allows us to decouple the effects of semantics and syntax on the sentence scores. All neural models and the bigram model perform best for canonical thematic role assignments. BERT and DistilBERT almost achieve an average AUC of 1 for these assignments, which is much higher than humans and than their own AUCs for the canonical case orders. To the contrary, the LSTM performance for these assignments is well below that of humans and below its own performances for the canonical case orders, suggesting that the LSTM is better at identifying acceptable syntactic structures than picking up semantic cues. 

\noindent Surprisingly, the Transformers perform much worse for semantic role assignments where the patient (dative case) is typically an inanimate object (lexical item 3). This drop in performance is also visible for the LSTM, albeit to a lesser extent. This shows that the BERT models specifically have learned a strong semantic constraint, on top of syntactic ones: $[DAT +HUMAN] >> [DAT -HUMAN]$, i.e dative NPs are more acceptable as human rather than
non-human entities. Together with the results from the markedness analysis, these findings show that the Transformer MLMs have learned more complex representations than the LSTM LM, combining syntactic and semantic generalizations that humans do not make.

\begin{table}[t]
\centering\small
\begin{tabular}{l @{\hspace{0.3em}} c @{\hspace{0.5em}}  c @{\hspace{0.5em}} c@{\hspace{0.5em}} c @{\hspace{0.5em}} c @{\hspace{0.5em}} c}
\toprule
Role &        &       &       &       &       & Distil-\\
assignment & humans & unigr.&  bigr.&  LSTM &  BERT &  BERT \\
\midrule
ag1,re2,pa3 & 0.79 & 0.52 & 0.58 & 0.66 & 0.96 & 0.90 \\
ag2,re1,pa3 & 0.79 & 0.42 & 0.54 & 0.63 & 0.95 & 0.87 \\
ag2,re3,pa2 & 0.58 & 0.60 & 0.52 & 0.52 & 0.67 & 0.60 \\
ag2,re3,pa1 & 0.56 & 0.51 & 0.47 & 0.48 & 0.67 & 0.61 \\
ag3,re1,pa2 & 0.55 & 0.48 & 0.47 & 0.56 & 0.83 & 0.74 \\
ag3,re2,pa1 & 0.54 & 0.50 & 0.46 & 0.57 & 0.83 & 0.76 \\
\it Avg &\it 0.63&\it 0.51&\it 0.51&\it 0.57&\it 0.82&\it 0.75\\
\bottomrule
\end{tabular}
\caption{(1-6) \mvss AUCs across humans and models averaged by role assignment and ranked by decreasing \textit{plausibility} (ag1, re2, pa3 most plausible).}
\label{tab:plausibility}
\vspace{-0.5em}
\end{table}

\subsection{(1-2) \mvss}
To assess the influence of case frequencies on human and model representations of verb argument structures, we restrict our analysis to (1-2) \mvss\footnote{Results for humans and models averaged by case order and semantic roles are shown in the Appendix}. For each input sentence, we limit the set of 6 minimally different unacceptable sentences to the 2 sentences with the same case violation. For instance, a (1-2) nominative \mvs contrasts one sentence experimentally labelled as acceptable with its 2 minimally different unacceptable sentences showcasing a doubled nominative. Table \ref{tab:1vs2} shows the average AUCs obtained by humans and models on these (1-2) \mvss and compares them to the average AUCs for (1-6) \mvss. 

\begin{table}[t]
\centering\small
\begin{tabular}{l @{\hspace{0.3em}} c @{\hspace{0.4em}}  c @{\hspace{0.4em}} c@{\hspace{0.4em}} c @{\hspace{0.4em}} c @{\hspace{0.4em}} c}
\toprule
Minimal &   &   &   &   &   & Distil-  \\
variation sets &  humans &  unigr. &  bigr. &  LSTM &  BERT &  BERT \\
\midrule
1-6 & 0.63 & 0.51 & 0.51 & 0.57 & 0.82 & 0.75 \\
1-2 nom & 0.59 &\it 0.10* &\it 0.12* &\it 0.39* & 0.76 & 0.71 \\
1-2 acc & 0.64 & 0.58 & 0.58 & 0.48 & 0.78 & 0.68 \\
1-2 dat & 0.68 & 0.84 & 0.81 & 0.84 & 0.91 & 0.85 \\
\bottomrule
\end{tabular}
\caption{AUCs across humans and models averaged by \mvs. *AUCs < 0.5 (chance level) mean that sentences containing a violation receive a higher score than grammatical sentences.}
\label{tab:1vs2}
\vspace{-1em}
\end{table}

We find that humans and models all achieve higher AUCs on (1-2) accusative \mvss than on (1-2) nominative \mvss, and even higher on (1-2) dative \mvss, to the exception of DistilBERT for (1-2) accusative \mvss. Humans and models tend to assign lower scores to sentences with doubled dative NPs, likely because these sentences lack either a nominative NP or an accusative NP, both of which are more frequent than dative NPs. Such behavior is probably due to the fact that German being a non pro-drop language, every verb must have a nominative case, making nominative more frequent than accusative, and dative even less common.

The frequency bias is worse for models directly based on frequency, such as our unigram and bigram models, and for the LSTM. The well below chance AUC of the LSTM for (1-2) nominative \mvss shows that the LSTM assigns a higher probability to sentences containing multiple nominative NPs than to correct sentences with only one nominative NP. In other words, the LSTM considers acceptable sentences \textit{less} acceptable than impossible sentences when the case violation is a double nominative NP. 

A case frequency analysis of the original German Wikipedia corpus using a spaCy German dependency tagger\footnote{de\_core\_news\_sm, trained on the TIGER and WikiNER corpora.} \cite{spacy2} supports this hypothesis: the tagger identifies 2x more nominative NPs than accusative NPs, and 8x more accusative NPs than dative NPs. The frequency bias of the  LSTM and \textit{n}-grams baselines directly reflects the frequency of nominative, accusative and dative in the language. Transformers are subject to this effect but to a lesser extent, and consistently achieve higher AUCs than humans for all (1-2) \mvss. 

\subsection{Correlations between human and model ratings}

In Table \ref{tab:correlations}, we show correlations between human acceptability judgments and model scores. All neural models are strongly correlated with humans while the unigram baseline and humans are negatively correlated. Unexpectedly, the LSTM and BERT show very similar correlation levels to humans, while the correlation between humans and DistilBERT is remarkably high. The origin of such a difference between BERT and DistilBERT is unclear. Understanding it would require to study how a parameter space reduction influences the representations learned by BERT models. 

\begin{table}[t]
\centering\small
\begin{tabular}{lc}
\toprule
 &  (1-6) minimal \\
Pearson correlation & variation sets \\
\midrule
Humans - unigrams &    -0.26 \\
Humans - bigrams &     0.58 \\
Humans - LSTM &     0.71 \\
Humans - BERT &     0.73 \\
Humans - DistilBERT &     0.81 \\
\bottomrule
\end{tabular}
\caption{Pearson correlation coefficients between humans and models.}
\label{tab:correlations}
\vspace{-1em}
\end{table}
\vspace{-0.5em}
\section{Discussion}
\vspace{-0.5em}

We set up a well controlled acceptability test for the processing of argument structures in humans, LSTM LMs and MLM BERT models. On average the LSTM achieve lower scores and the Transformers higher scores than the gold standard defined by human AUCs. The Transformers' scores are substantially higher than chance not only for human-established canonical case orders and thematic roles, but also for permutations where scores given by humans are moderate or close to chance. The LSTM's comparatively poor performance may be partly due to its smaller training data and number of parameters. Vocabulary coverage, however, should be ruled out as a possible issue since all tokens, including case variations, were present in its vocabulary.

Our results also show that the LSTM, unlike the Transformers, is biased by frequency cues present in the training dataset. First, the LSTM is overly sensitive to NPs' frequency distribution. It is terrible at discerning acceptable sentences from unacceptable sentences when the case violation involves a doubled nominative, the most frequent case in the training dataset. This is not the case for the Transformers, which perform strongly regardless of the case violation.

Second, the LSTM performs well below chance for the most marked case orders. These argument structures, in practice rarely or never encountered in the language, are considered unacceptable. In constrast, the Transformers perform consistently well for all case orders, even those likely to be absent from the training dataset, such as ADN. German BERT in particular is able to generalize grammatical constraints, e.g doubled cases are strongly unacceptable, beyond case order frequencies. We suspect that it might even overgeneralize them, neglecting language usage.

Finally, we find that grammatical constraints previously identified in humans also generally control our LMs' representations, although the  German BERT have learned a peculiar semantic rule absent from other models. Overall, our results suggest that the Transformer MLMs, unlike the LSTM LM, learn syntactic and semantic generalizations that may go beyond those made by humans.

\section*{Acknowledgments}

The team's project is funded by the European Research Council (ERC-2011-AdG-295810 BOOTPHON), the Agence Nationale pour la Recherche (ANR-10-LABX-0087 IEC, ANR-10-IDEX-0001-02 PSL*), Almerys (industrial chair Data Science and Security), and grants from Facebook AI Research (Research Grant), Google (Faculty Research Award), Microsoft Research (Azure Credits and Grant), and Amazon Web Service (AWS Research Credits).

\bibliography{references}

\begin{thebibliography}{27}
\expandafter\ifx\csname natexlab\endcsname\relax\def\natexlab#1{#1}\fi

\bibitem[{Allen and Seidenberg(1999)}]{allen1999:emergence}
Joseph Allen and Mark~S Seidenberg. 1999.
\newblock The emergence of grammaticality in connectionist networks.
\newblock \emph{The emergence of language}, pages 115--151.

\bibitem[{Devlin et~al.(2019)Devlin, Chang, Lee, and
  Toutanova}]{devlin2018bert}
Jacob Devlin, Ming-Wei Chang, Kenton Lee, and Kristina Toutanova. 2019.
\newblock \href {https://www.aclweb.org/anthology/N19-1423.pdf} {{BERT}:
  Pre-training of deep bidirectional transformers for language understanding}.
\newblock In \emph{Proceedings of the 2019 Conference of the North American
  Chapter of the Association for Computational Linguistics}, pages 4171--4186.

\bibitem[{Futrell et~al.(2019)Futrell, Wilcox, Morita, Qian, Ballesteros, and
  Levy}]{futrell:2019}
Richard Futrell, Ethan Wilcox, Takashi Morita, Peng Qian, Miguel Ballesteros,
  and Roger Levy. 2019.
\newblock \href {https://www.aclweb.org/anthology/N19-1004.pdf} {Neural
  language models as psycholinguistic subjects: Representations of syntactic
  state}.
\newblock In \emph{Proceedings of the 2019 Conference of the North American
  Chapter of the Association for Computational Linguistics}, pages 32--42.

\bibitem[{Goldberg(2019)}]{goldberg:2019}
Yoav Goldberg. 2019.
\newblock \href {http://arxiv.org/abs/1901.05287} {Assessing bert's syntactic
  abilities}.
\newblock \emph{arXiv preprint 1901.05287}.

\bibitem[{Gulordava et~al.(2018)Gulordava, Bojanowski, Grave, Linzen, and
  Baroni}]{gulordava:2018}
Kristina Gulordava, Piotr Bojanowski, Edouard Grave, Tal Linzen, and Marco
  Baroni. 2018.
\newblock \href {https://www.aclweb.org/anthology/N18-1108} {Colorless green
  recurrent networks dream hierarchically}.
\newblock In \emph{Proceedings of the 2019 Conference of the North American
  Chapter of the Association for Computational Linguistics}, pages 1195--1205.

\bibitem[{Hahn and Baroni(2019)}]{hahn2019tabula}
Michael Hahn and Marco Baroni. 2019.
\newblock \href {https://arxiv.org/abs/1906.07285} {Tabula nearly rasa: Probing
  the linguistic knowledge of character-level neural language models trained on
  unsegmented text}.
\newblock In \emph{Transactions of the Association for Computational
  Linguistics}, pages 467--484.

\bibitem[{Hochreiter and Schmidhuber(1997)}]{hochreiter1997long}
Sepp Hochreiter and J{\"u}rgen Schmidhuber. 1997.
\newblock \href {https://doi.org/10.1162/neco.1997.9.8.1735} {Long short-term
  memory}.
\newblock In \emph{Neural computation}, volume~9, pages 1735--1780. MIT Press.

\bibitem[{Honnibal and Montani(2017)}]{spacy2}
Matthew Honnibal and Ines Montani. 2017.
\newblock \href {https://spacy.io/models} {{spaCy 2}: Natural language
  understanding with {B}loom embeddings, convolutional neural networks and
  incremental parsing}.
\newblock Unpublished.

\bibitem[{Keller(2001)}]{Keller2001Gradience}
Frank Keller. 2001.
\newblock \href {http://roa.rutgers.edu/files/677-0804/677-KELLER-0-0.PDF}
  {\emph{Gradience in grammar : experimental and computational aspects of
  degrees of grammaticality}}.
\newblock Ph.D. thesis, University of Edinburgh.

\bibitem[{Lakretz et~al.(2019)Lakretz, Kruszewski, Desbordes, Hupkes, Dehaene,
  and Baroni}]{Lakretz2019TheEO}
Yair Lakretz, Germ{\'a}n Kruszewski, Theo Desbordes, Dieuwke Hupkes, Stanislas
  Dehaene, and Marco Baroni. 2019.
\newblock \href {https://www.aclweb.org/anthology/N19-1002.pdf} {The emergence
  of number and syntax units in lstm language models}.
\newblock In \emph{Proceedings of the 2019 Conference of the North American
  Chapter of the Association for Computational Linguistics}, pages 11--20.

\bibitem[{Linzen et~al.(2016)Linzen, Dupoux, and Goldberg}]{linzen:2016}
Tal Linzen, Emmanuel Dupoux, and Yoav Goldberg. 2016.
\newblock \href {https://www.aclweb.org/anthology/Q16-1037/} {Assessing the
  ability of lstms to learn syntax-sensitive dependencies}.
\newblock In \emph{Trans. Assoc. Comput. Linguistics}, volume~4, pages
  521--535.

\bibitem[{Marvin and Linzen(2018)}]{marvin-linzen:2018}
Rebecca Marvin and Tal Linzen. 2018.
\newblock \href {https://www.aclweb.org/anthology/D18-1151} {Targeted syntactic
  evaluation of language models}.
\newblock In \emph{Proceedings of the 2018 Conference on Empirical Methods in
  Natural Language Processing}, pages 1192--1202.

\bibitem[{Peters et~al.(2018)Peters, Neumann, Iyyer, Gardner, Clark, Lee, and
  Zettlemoyer}]{elmo}
Matthew Peters, Mark Neumann, Mohit Iyyer, Matt Gardner, Christopher Clark,
  Kenton Lee, and Luke Zettlemoyer. 2018.
\newblock \href {https://www.aclweb.org/anthology/N18-1202/} {Deep
  contextualized word representations}.
\newblock In \emph{Proceedings of the 2018 Conference of the North American
  Chapter of the Association for Computational Linguistics}, pages 2227--2237.

\bibitem[{Raffel et~al.(2020)Raffel, Shazeer, Roberts, Lee, Narang, Matena,
  Zhou, Li, and Liu}]{raffel2020T5}
Colin Raffel, Noam Shazeer, Adam Roberts, Katherine Lee, Sharan Narang, Michael
  Matena, Yanqi Zhou, Wei Li, and Peter~J. Liu. 2020.
\newblock \href {http://jmlr.org/papers/v21/20-074.html} {Exploring the limits
  of transfer learning with a unified text-to-text transformer}.
\newblock \emph{Journal of Machine Learning Research}, 21(140):1--67.

\bibitem[{Ravfogel et~al.(2018)Ravfogel, Tyers, and Goldberg}]{ravfogel2018can}
Shauli Ravfogel, Francis~M Tyers, and Yoav Goldberg. 2018.
\newblock \href {https://www.aclweb.org/anthology/W18-5412/} {Can {LSTM} learn
  to capture agreement? the case of basque}.
\newblock In \emph{Proceedings of the Workshop: Analyzing and Interpreting
  Neural Networks for NLP}, pages 98--107.

\bibitem[{Salazar et~al.(2020)Salazar, Liang, Nguyen, and
  Kirchhoff}]{salazar2020masked}
Julian Salazar, Davis Liang, Toan~Q. Nguyen, and Katrin Kirchhoff. 2020.
\newblock \href {https://www.aclweb.org/anthology/2020.acl-main.240} {Masked
  language model scoring}.
\newblock In \emph{Proceedings of the 58th Annual Meeting of the Association
  for Computational Linguistics}, pages 2699--2712.

\bibitem[{Sanh et~al.(2019)Sanh, Debut, Chaumond, and
  Wolf}]{sanh2019distilbert}
Victor Sanh, Lysandre Debut, Julien Chaumond, and Thomas Wolf. 2019.
\newblock \href {http://arxiv.org/abs/1910.01108} {Distilbert, a distilled
  version of bert: smaller, faster, cheaper and lighter}.
\newblock In \emph{NeurIPS $EMC^2$ Workshop}.

\bibitem[{Shin et~al.(2019)Shin, Lee, and Jung}]{Shin2019EffectiveSS}
Joonbo Shin, Yoonhyung Lee, and Kyomin Jung. 2019.
\newblock \href {http://proceedings.mlr.press/v101/shin19a.html} {Effective
  sentence scoring method using bert for speech recognition}.
\newblock In \emph{Proceedings of The 11th Asian Conference on Machine
  Learning}, pages 1081--1093.

\bibitem[{Tran et~al.(2018)Tran, Bisazza, and Monz}]{tran:2018}
Ke~Tran, Arianna Bisazza, and Christof Monz. 2018.
\newblock \href {https://www.aclweb.org/anthology/D18-1503.pdf} {The importance
  of being recurrent for modeling hierarchical structure}.
\newblock In \emph{Proceedings of the 2018 Conference on Empirical Methods in
  Natural Language Processing}, pages 4731--4736.

\bibitem[{Uszkoreit(1987)}]{Uszkoreit1987Word}
Hans Uszkoreit. 1987.
\newblock \emph{Word Order and Constituent Structure in German}, volume~8 of
  \emph{{CSLI} Lecture Notes}.
\newblock {CSLI} Publications, Stanford, {CA}.

\bibitem[{Vaswani et~al.(2017)Vaswani, Shazeer, Parmar, Uszkoreit, Jones,
  Gomez, Kaiser, and Polosukhin}]{vaswani2017attention}
Ashish Vaswani, Noam Shazeer, Niki Parmar, Jakob Uszkoreit, Llion Jones,
  Aidan~N. Gomez, undefinedukasz Kaiser, and Illia Polosukhin. 2017.
\newblock Attention is all you need.
\newblock In \emph{Proceedings of the 31st International Conference on Neural
  Information Processing Systems}, pages 5998--6008.

\bibitem[{Wang and Cho(2019)}]{Wang2019BERTHA}
Alex Wang and Kyunghyun Cho. 2019.
\newblock \href {http://arxiv.org/abs/1902.04094} {Bert has a mouth, and it
  must speak: Bert as a markov random field language model}.
\newblock \emph{arXiv preprint 1902.04094}.

\bibitem[{Warstadt and Bowman(2019)}]{cola_grammatical}
Alex Warstadt and Samuel~R Bowman. 2019.
\newblock \href {https://arxiv.org/abs/1901.03438} {Grammatical analysis of
  pretrained sentence encoders with acceptability judgments}.
\newblock \emph{arXiv preprint 1901.03438}.

\bibitem[{Warstadt et~al.(2019{\natexlab{a}})Warstadt, Cao, Grosu, Peng, Blix,
  Nie, Alsop, Bordia, Liu, Parrish, Wang, Phang, Mohananey, Htut, Jeretič, and
  Bowman}]{warstadt2019investigating}
Alex Warstadt, Yu~Cao, Ioana Grosu, Wei Peng, Hagen Blix, Yining Nie, Anna
  Alsop, Shikha Bordia, Haokun Liu, Alicia Parrish, Sheng-Fu Wang, Jason Phang,
  Anhad Mohananey, Phu~Mon Htut, Paloma Jeretič, and Samuel~R. Bowman.
  2019{\natexlab{a}}.
\newblock \href {https://www.aclweb.org/anthology/D19-1286.pdf} {Investigating
  bert's knowledge of language: Five analysis methods with npis}.
\newblock In \emph{Proceedings of the 2019 Conference on Empirical Methods in
  Natural Language Processing and the 9th International Joint Conference on
  Natural Language Processing}, pages 2877--2887.

\bibitem[{Warstadt et~al.(2019{\natexlab{b}})Warstadt, Singh, and
  Bowman}]{cola}
Alex Warstadt, Amanpreet Singh, and Samuel~R Bowman. 2019{\natexlab{b}}.
\newblock \href {https://transacl.org/ojs/index.php/tacl/article/view/1710}
  {Neural network acceptability judgments}.
\newblock \emph{Trans. Assoc. Comput. Linguistics}, 7:625--641.

\bibitem[{Wilcox et~al.(2018)Wilcox, Levy, Morita, and Futrell}]{wilcox:2018}
Ethan Wilcox, Roger~P. Levy, Takashi Morita, and Richard Futrell. 2018.
\newblock \href {https://www.aclweb.org/anthology/W18-5423/} {What do {RNN}
  language models learn about filler–gap dependencies?}
\newblock In \emph{Workshop on Analyzing and Interpreting Neural Networks for
  NLP}.

\bibitem[{Wilcox et~al.(2019)Wilcox, Qian, Futrell, Ballesteros, and
  Levy}]{wilcox2019structural}
Ethan Wilcox, Peng Qian, Richard Futrell, Miguel Ballesteros, and Roger Levy.
  2019.
\newblock \href {https://www.aclweb.org/anthology/N19-1334.pdf} {Structural
  supervision improves learning of non-local grammatical dependencies}.
\newblock In \emph{Proceedings of the 2019 Conference of the North American
  Chapter of the Association for Computational Linguistics}, pages 3302--3312.

\end{thebibliography}
\bibliographystyle{acl_natbib}




\newpage
\appendix

\section{Models AUCs for (1-6) \mvss}
\label{sec:1-6_models}
In tables \ref{tab:unigram_1to6}, \ref{tab:bigram_1to6}, \ref{tab:lstm_1to6}, \ref{tab:bert_1to6} and \ref{tab:distilbert_1to6}, we report AUCs for our (1-6) \mvs across all models. AUCs are averaged by case order and ranked by increasing \textit{markedness} (NDA least marked) and decreasing \textit{plausibility} (ag1,re2,pa3 most plausible) according to humans.
\begin{table*}[t]
\centering
\begin{tabular}{lccccccc}
\toprule
Role assignment / & ag1, re2, & ag2, re1& ag1, re3, & ag2, re3, & ag3, re1, & ag3, re2, & \it Avg \\
Case order &  pa3 & pa3  & pa2 & pa1 & pa2 & pa1 & \it markedness\\
\midrule
NDA & 0.57 & 0.44 & 0.60 & 0.50 & 0.47 & 0.49 & \it0.51 \\
NAD & 0.49 & 0.40 & 0.58 & 0.48 & 0.47 & 0.47 & \it0.48 \\
DNA & 0.57 & 0.44 & 0.60 & 0.50 & 0.47 & 0.49 & \it0.51 \\
AND & 0.49 & 0.40 & 0.58 & 0.48 & 0.47 & 0.47 & \it0.48 \\
DAN & 0.51 & 0.42 & 0.62 & 0.55 & 0.51 & 0.53 & \it0.52 \\
ADN & 0.51 & 0.42 & 0.62 & 0.55 & 0.51 & 0.53 & \it0.52 \\
\it Avg plaus. & \it0.52 & \it0.42 & \it0.60 & \it0.51 & \it0.48 & \it0.50 & \it0.51 \\
\bottomrule
\end{tabular}
\caption{(1-6) \mvss unigram AUCs (higher is better) averaged by case order and ranked by increasing \textit{markedness} (NDA least marked) and decreasing \textit{plausibility} (ag1,re2,pa3 most plausible).}
\label{tab:unigram_1to6}
\end{table*}

\begin{table*}[t]
\centering
\begin{tabular}{lccccccc}
\toprule
Role assignment / & ag1, re2, & ag2, re1& ag1, re3, & ag2, re3, & ag3, re1, & ag3, re2, & \it Avg \\
Case order &  pa3 & pa3  & pa2 & pa1 & pa2 & pa1 & \it markedness\\
\midrule
NDA & 0.65 & 0.63 & 0.48 & 0.45 & 0.52 & 0.51 & \it0.54 \\
NAD & 0.56 & 0.53 & 0.58 & 0.54 & 0.48 & 0.47 & \it0.53 \\
DNA & 0.68 & 0.59 & 0.55 & 0.48 & 0.41 & 0.37 & \it0.51 \\
AND & 0.49 & 0.46 & 0.56 & 0.51 & 0.33 & 0.36 & \it0.45 \\
DAN & 0.64 & 0.63 & 0.57 & 0.53 & 0.63 & 0.64 & \it0.61 \\
ADN & 0.43 & 0.41 & 0.38 & 0.33 & 0.43 & 0.38 & \it0.39 \\
\it Avg plaus. & \it0.58 & \it0.54 & \it0.52 & \it0.47 & \it0.47 & \it0.46 & \it0.51 \\
\bottomrule
\end{tabular}
\caption{(1-6) \mvss bigram AUCs (higher is better) averaged by case order and ranked by increasing \textit{markedness} (NDA least marked) and decreasing \textit{plausibility} (ag1,re2,pa3 most plausible).}
\label{tab:bigram_1to6}
\end{table*}

\begin{table*}[t]
\centering
\begin{tabular}{lccccccc}
\toprule
Role assignment / & ag1, re2, & ag2, re1& ag1, re3, & ag2, re3, & ag3, re1, & ag3, re2, & \it Avg \\
Case order &  pa3 & pa3  & pa2 & pa1 & pa2 & pa1 & \it markedness\\
\midrule
NDA & 0.91 & 0.89 & 0.67 & 0.60 & 0.78 & 0.79 & \it0.77 \\
NAD & 0.80 & 0.73 & 0.79 & 0.73 & 0.70 & 0.73 & \it0.75 \\
DNA & 0.79 & 0.79 & 0.53 & 0.51 & 0.48 & 0.51 & \it0.60 \\
AND & 0.43 & 0.41 & 0.45 & 0.42 & 0.33 & 0.32 & \it0.39 \\
DAN & 0.62 & 0.61 & 0.44 & 0.42 & 0.64 & 0.64 & \it0.56 \\
ADN & 0.40 & 0.37 & 0.22 & 0.22 & 0.43 & 0.44 & \it0.35 \\
\it Avg plaus. & \it0.66 & \it0.63 & \it0.52 & \it0.48 & \it0.56 & \it0.57 & \it0.57 \\
\bottomrule
\end{tabular}
\caption{(1-6) \mvss LSTM AUCs (higher is better) averaged by case order and ranked by increasing \textit{markedness} (NDA least marked) and decreasing \textit{plausibility} (ag1,re2,pa3 most plausible).}
\label{tab:lstm_1to6}
\end{table*}

\begin{table*}[t]
\centering
\begin{tabular}{lccccccc}
\toprule
Role assignment / & ag1, re2, & ag2, re1& ag1, re3, & ag2, re3, & ag3, re1, & ag3, re2, & \it Avg \\
Case order &  pa3 & pa3  & pa2 & pa1 & pa2 & pa1 & \it markedness\\
\midrule
NDA & 0.99 & 0.99 & 0.57 & 0.57 & 0.87 & 0.86 & \it0.81 \\
NAD & 0.95 & 0.96 & 0.87 & 0.86 & 0.88 & 0.91 & \it0.90 \\
DNA & 0.98 & 0.99 & 0.66 & 0.62 & 0.83 & 0.81 & \it0.82 \\
AND & 0.94 & 0.94 & 0.74 & 0.78 & 0.68 & 0.71 & \it0.80 \\
DAN & 0.93 & 0.89 & 0.58 & 0.61 & 0.84 & 0.82 & \it0.78 \\
ADN & 0.95 & 0.95 & 0.58 & 0.59 & 0.89 & 0.87 & \it0.81 \\
\it Avg plaus. & \it0.96 & \it0.95 & \it0.67 & \it0.67 & \it0.83 & \it0.83 & \it0.82 \\
\bottomrule
\end{tabular}
\caption{(1-6) \mvss BERT AUCs (higher is better) averaged by case order and ranked by increasing \textit{markedness} (NDA least marked) and decreasing \textit{plausibility} (ag1,re2,pa3 most plausible).}
\label{tab:bert_1to6}
\end{table*}

\begin{table*}[t]
\centering
\begin{tabular}{lccccccc}
\toprule
Role assignment / & ag1, re2, & ag2, re1& ag1, re3, & ag2, re3, & ag3, re1, & ag3, re2, & \it Avg \\
Case order &  pa3 & pa3  & pa2 & pa1 & pa2 & pa1 & \it markedness\\
\midrule
NDA & 0.99 & 0.98 & 0.63 & 0.62 & 0.83 & 0.84 & \it0.81 \\
NAD & 0.96 & 0.94 & 0.83 & 0.85 & 0.85 & 0.89 & \it0.89 \\
DNA & 0.96 & 0.95 & 0.62 & 0.59 & 0.75 & 0.69 & \it0.76 \\
AND & 0.92 & 0.86 & 0.68 & 0.64 & 0.61 & 0.66 & \it0.73 \\
DAN  & 0.68 & 0.64 & 0.43 & 0.48 & 0.65 & 0.71 & \it0.60 \\
ADN  & 0.88 & 0.87 & 0.42 & 0.45 & 0.77 & 0.79 & \it0.70 \\
\it Avg plaus. & \it0.90 & \it0.87 & \it0.60 & \it0.61 & \it0.74 & \it0.76 & \it0.75 \\
\bottomrule
\end{tabular}
\caption{(1-6) \mvss DistilBERT AUCs (higher is better) averaged by case order and ranked by increasing \textit{markedness} (NDA least marked) and decreasing \textit{plausibility} (ag1,re2,pa3 most plausible).}
\label{tab:distilbert_1to6}
\end{table*}

\section{Models AUCs (1-2) nominative \mvss}
\label{sec:1-2_nom_tables}
In tables \ref{tab:unigram_1to2_nom}, \ref{tab:bigram_1to2_nom}, \ref{tab:lstm_1to2_nom}, \ref{tab:bert_1to2_nom} and \ref{tab:distilbert_1to2_nom}, we show AUCs for our (1-2) nominative \mvs across all models. AUCs are averaged by case order and ranked by increasing \textit{markedness} (NDA least marked) and decreasing \textit{plausibility} (ag1,re2,pa3 most plausible) according to humans.

\begin{table*}[t]
\centering
\begin{tabular}{lccccccc}
\toprule
Role assignment / & ag1, re2, & ag2, re1& ag1, re3, & ag2, re3, & ag3, re1, & ag3, re2, & \it Avg \\
Case order &  pa3 & pa3  & pa2 & pa1 & pa2 & pa1 & \it markedness\\
\midrule
NDA & 0.99 & 1.00 & 0.59 & 0.55 & 0.59 & 0.58 & \it0.72 \\
NAD & 0.95 & 0.91 & 0.76 & 0.79 & 0.60 & 0.62 & \it0.77 \\
DNA & 0.83 & 0.83 & 0.44 & 0.36 & 0.39 & 0.40 & \it0.54 \\
AND & 0.61 & 0.73 & 0.66 & 0.58 & 0.43 & 0.45 & \it0.58 \\
DAN & 0.51 & 0.44 & 0.31 & 0.37 & 0.50 & 0.55 & \it0.45 \\
ADN & 0.59 & 0.57 & 0.34 & 0.28 & 0.56 & 0.53 & \it0.48 \\
\it Avg plaus. & \it0.75 & \it0.75 & \it0.52 & \it0.49 & \it0.51 & \it0.52 & \it0.59 \\
\bottomrule
\end{tabular}
\caption{(1-2) nominative \mvss human AUCs (higher is better) averaged by case order and ranked by increasing \textit{markedness} (NDA least marked) and decreasing \textit{plausibility} (ag1,re2,pa3 most plausible).}
\label{tab:human_1to2_nom}
\end{table*}

\begin{table*}[t]
\centering
\begin{tabular}{lccccccc}
\toprule
Role assignment / & ag1, re2, & ag2, re1& ag1, re3, & ag2, re3, & ag3, re1, & ag3, re2, & \it Avg \\
Case order &  pa3 & pa3  & pa2 & pa1 & pa2 & pa1 & \it markedness\\
\midrule
NDA & 0.15 & 0.00 & 0.09 & 0.09 & 0.01 & 0.15 & \it0.08 \\
NAD & 0.04 & 0.04 & 0.13 & 0.02 & 0.14 & 0.01 & \it0.06 \\
DNA & 0.15 & 0.00 & 0.09 & 0.09 & 0.01 & 0.15 & \it0.08 \\
AND & 0.04 & 0.04 & 0.13 & 0.02 & 0.14 & 0.01 & \it0.06 \\
DAN & 0.19 & 0.08 & 0.21 & 0.13 & 0.15 & 0.17 & \it0.15 \\
ADN & 0.19 & 0.08 & 0.21 & 0.13 & 0.15 & 0.17 & \it0.15 \\
\it Avg plaus. & \it0.13 & \it0.04 & \it0.14 & \it0.08 & \it0.10 & \it0.11 & \it0.10 \\
\bottomrule
\end{tabular}
\caption{(1-2) nominative \mvss unigram AUCs (higher is better) averaged by case order and ranked by increasing \textit{markedness} (NDA least marked) and decreasing \textit{plausibility} (ag1,re2,pa3 most plausible).}
\label{tab:unigram_1to2_nom}
\end{table*}

\begin{table*}[t]
\centering
\begin{tabular}{lccccccc}
\toprule
Role assignment / & ag1, re2, & ag2, re1& ag1, re3, & ag2, re3, & ag3, re1, & ag3, re2, & \it Avg \\
Case order &  pa3 & pa3  & pa2 & pa1 & pa2 & pa1 & \it markedness\\
\midrule
NDA & 0.25 & 0.15 & 0.14 & 0.08 & 0.09 & 0.14 & \it0.14 \\
NAD & 0.22 & 0.18 & 0.10 & 0.06 & 0.05 & 0.05 & \it0.11 \\
DNA & 0.20 & 0.15 & 0.12 & 0.12 & 0.03 & 0.08 & \it0.12 \\
AND & 0.04 & 0.05 & 0.09 & 0.07 & 0.04 & 0.01 & \it0.05 \\
DAN & 0.24 & 0.23 & 0.17 & 0.12 & 0.17 & 0.22 & \it0.19 \\
ADN & 0.14 & 0.11 & 0.12 & 0.07 & 0.11 & 0.11 &  \it0.11 \\
\it Avg plaus. & \it0.18 & \it0.14 & \it0.12 & \it0.09 & \it0.08 & \it0.10 & \it0.12 \\
\bottomrule
\end{tabular}
\caption{(1-2) nominative \mvss bigram AUCs (higher is better) averaged by case order and ranked by increasing \textit{markedness} (NDA least marked) and decreasing \textit{plausibility} (ag1,re2,pa3 most plausible).}
\label{tab:bigram_1to2_nom}
\end{table*}

\begin{table*}[t]
\centering
\begin{tabular}{lccccccc}
\toprule
Role assignment / & ag1, re2, & ag2, re1& ag1, re3, & ag2, re3, & ag3, re1, & ag3, re2, & \it Avg \\
Case order &  pa3 & pa3  & pa2 & pa1 & pa2 & pa1 & \it markedness\\
\midrule
NDA & 0.79 & 0.78 & 0.59 & 0.57 & 0.69 & 0.77 & \it0.70 \\
NAD & 0.78 & 0.73 & 0.57 & 0.57 & 0.60 & 0.63 & \it0.65 \\
DNA & 0.48 & 0.49 & 0.22 & 0.23 & 0.25 & 0.35 & \it0.34 \\
AND & 0.14 & 0.11 & 0.23 & 0.19 & 0.13 & 0.10 & \it0.15 \\
DAN & 0.36 & 0.34 & 0.15 & 0.13 & 0.42 & 0.36 & \it0.29 \\
ADN & 0.15 & 0.14 & 0.12 & 0.08 & 0.31 & 0.33 & \it0.19 \\
\it Avg plaus. & \it0.45 & \it0.43 & \it0.31 & \it0.29 & \it0.40 & \it0.42 & \it0.39 \\
\bottomrule
\end{tabular}
\caption{(1-2) nominative \mvss LSTM AUCs (higher is better) averaged by case order and ranked by increasing \textit{markedness} (NDA least marked) and decreasing \textit{plausibility} (ag1,re2,pa3 most plausible).}
\label{tab:lstm_1to2_nom}
\end{table*}

\begin{table*}[t]
\centering
\begin{tabular}{lccccccc}
\toprule
Role assignment / & ag1, re2, & ag2, re1& ag1, re3, & ag2, re3, & ag3, re1, & ag3, re2, & \it Avg \\
Case order &  pa3 & pa3  & pa2 & pa1 & pa2 & pa1 & \it markedness\\
\midrule
NDA & 1.00 & 1.00 & 0.40 & 0.45 & 0.89 & 0.88 & \it0.77 \\
NAD & 0.96 & 0.96 & 0.83 & 0.82 & 0.87 & 0.94 & \it0.90 \\
DNA & 0.97 & 1.00 & 0.51 & 0.54 & 0.82 & 0.83 & \it0.78 \\
AND & 0.93 & 0.90 & 0.67 & 0.77 & 0.61 & 0.64 & \it0.75 \\
DAN & 0.88 & 0.81 & 0.49 & 0.51 & 0.76 & 0.73 & \it0.70 \\
ADN & 0.88 & 0.89 & 0.33 & 0.37 & 0.81 & 0.76 & \it0.67 \\
\it Avg plaus. & \it0.94 & \it0.93 & \it0.54 & \it0.58 & \it0.79 & \it0.80 & \it0.76 \\
\bottomrule
\end{tabular}
\caption{(1-2) nominative \mvss BERT AUCs (higher is better) averaged by case order and ranked by increasing \textit{markedness} (NDA least marked) and decreasing \textit{plausibility} (ag1,re2,pa3 most plausible).}
\label{tab:bert_1to2_nom}
\end{table*}

\begin{table*}[t]
\centering
\begin{tabular}{lccccccc}
\toprule
Role assignment / & ag1, re2, & ag2, re1& ag1, re3, & ag2, re3, & ag3, re1, & ag3, re2, & \it Avg \\
Case order &  pa3 & pa3  & pa2 & pa1 & pa2 & pa1 & \it markedness\\
\midrule
NDA & 0.99 & 1.00 & 0.66 & 0.70 & 0.81 & 0.85 & \it0.83 \\
NAD & 0.95 & 0.97 & 0.78 & 0.79 & 0.85 & 0.87 & \it0.87 \\
DNA & 0.95 & 0.96 & 0.49 & 0.53 & 0.79 & 0.76 & \it0.75 \\
AND & 0.85 & 0.80 & 0.60 & 0.55 & 0.63 & 0.69 & \it0.69 \\
DAN & 0.63 & 0.61 & 0.41 & 0.42 & 0.59 & 0.60 & \it0.54 \\
ADN & 0.83 & 0.83 & 0.31 & 0.25 & 0.66 & 0.72 & \it0.60 \\
\it Avg plaus. & \it0.87 & \it0.86 & \it0.54 & \it0.54 & \it0.72 & \it0.75 & \it0.71 \\
\bottomrule
\end{tabular}
\caption{(1-2) nominative \mvss DistilBERT AUCs (higher is better) averaged by case order and ranked by increasing \textit{markedness} (NDA least marked) and decreasing \textit{plausibility} (ag1,re2,pa3 most plausible).}
\label{tab:distilbert_1to2_nom}
\end{table*}

\section{Models AUCs (1-2) accusative \mvss}
\label{sec:1-2_acc_tables}
Tables \ref{tab:unigram_1to2_acc}, \ref{tab:bigram_1to2_acc}, \ref{tab:lstm_1to2_acc}, \ref{tab:bert_1to2_acc} and \ref{tab:distilbert_1to2_acc} show AUCs for (1-2) accusative \mvs across all models. AUCs are averaged by case order and ranked by increasing \textit{markedness} (NDA least marked) and decreasing \textit{plausibility} (ag1,re2,pa3 most plausible) according to humans.

\begin{table*}[t]
\centering
\begin{tabular}{lccccccc}
\toprule
Role assignment / & ag1, re2, & ag2, re1& ag1, re3, & ag2, re3, & ag3, re1, & ag3, re2, & \it Avg \\
Case order &  pa3 & pa3  & pa2 & pa1 & pa2 & pa1 & \it markedness\\
\midrule
NDA & 0.99 & 0.98 & 0.57 & 0.56 & 0.55 & 0.53 & \it0.70 \\
NAD & 0.92 & 0.79 & 0.92 & 0.84 & 0.51 & 0.52 & \it0.75 \\
DNA & 0.87 & 0.88 & 0.50 & 0.40 & 0.63 & 0.38 & \it0.61 \\
AND & 0.74 & 0.71 & 0.63 & 0.64 & 0.43 & 0.49 & \it0.61 \\
DAN & 0.68 & 0.74 & 0.36 & 0.48 & 0.60 & 0.64 & \it0.58 \\
ADN & 0.69 & 0.71 & 0.45 & 0.41 & 0.57 & 0.62 & \it0.57 \\
\it Avg plaus. & \it0.82 & \it0.80 & \it0.57 & \it0.56 & \it0.55 & \it0.53 & \it0.64 \\
\bottomrule
\end{tabular}
\caption{(1-2) accusative \mvss human AUCs (higher is better) averaged by case order and ranked by increasing \textit{markedness} (NDA least marked) and decreasing \textit{plausibility} (ag1,re2,pa3 most plausible).}
\label{tab:human_1to2_acc}
\end{table*}

\begin{table*}[t]
\centering
\begin{tabular}{lccccccc}
\toprule
Role assignment / & ag1, re2, & ag2, re1& ag1, re3, & ag2, re3, & ag3, re1, & ag3, re2, & \it Avg \\
Case order &  pa3 & pa3  & pa2 & pa1 & pa2 & pa1 & \it markedness\\
\midrule
NDA & 0.15 & 0.00 & 0.09 &0.09 & 0.01 & 0.15 & \it0.08 \\
NAD & 0.04 & 0.04 & 0.13 & 0.02 & 0.14 & 0.01 & \it0.06 \\
DNA & 0.15 & 0.00 & 0.09 & 0.09 & 0.01 & 0.15 & \it0.08 \\
AND & 0.04 & 0.04 & 0.13 & 0.02 & 0.14 & 0.01 & \it0.06 \\
DAN & 0.19 & 0.08 & 0.21 & 0.13 & 0.15 & 0.17 & \it0.15 \\
ADN & 0.19 & 0.08 & 0.21 & 0.13 & 0.15 & 0.17 & \it0.15 \\
\it Avg plaus. & \it0.13 & \it0.04 & \it0.14 & \it0.08 & \it0.10 & \it0.11 & \it0.10 \\
\bottomrule
\end{tabular}
\caption{(1-2) accusative \mvss unigram AUCs (higher is better) averaged by case order and ranked by increasing \textit{markedness} (NDA least marked) and decreasing \textit{plausibility} (ag1,re2,pa3 most plausible).}
\label{tab:unigram_1to2_acc}
\end{table*}

\begin{table*}[t]
\centering
\begin{tabular}{lccccccc}
\toprule
Role assignment / & ag1, re2, & ag2, re1& ag1, re3, & ag2, re3, & ag3, re1, & ag3, re2, & \it Avg \\
Case order &  pa3 & pa3  & pa2 & pa1 & pa2 & pa1 & \it markedness\\
\midrule
NDA & 0.78 & 0.78 & 0.47 & 0.46 & 0.59 & 0.57 & \it0.61 \\
NAD & 0.61 & 0.54 & 0.72 & 0.69 & 0.55 & 0.56 & \it0.61 \\
DNA & 0.87 & 0.75 & 0.65 & 0.52 & 0.44 & 0.29 & \it0.59 \\
AND & 0.66 & 0.61 & 0.71 & 0.65 & 0.39 & 0.44 & \it0.58 \\
DAN & 0.70 & 0.70 & 0.58 & 0.54 & 0.80 & 0.76 & \it0.68 \\
ADN & 0.44 & 0.43 & 0.38 & 0.30 & 0.48 & 0.43 & \it0.41 \\
\it Avg plaus. & \it0.68 & \it0.64 & \it0.58 & \it0.53 & \it0.54 & \it0.51 & \it0.58 \\
\bottomrule
\end{tabular}
\caption{(1-2) accusative \mvss bigram AUCs (higher is better) averaged by case order and ranked by increasing \textit{markedness} (NDA least marked) and decreasing \textit{plausibility} (ag1,re2,pa3 most plausible).}
\label{tab:bigram_1to2_acc}
\end{table*}

\begin{table*}[t]
\centering
\begin{tabular}{lccccccc}
\toprule
Role assignment / & ag1, re2, & ag2, re1& ag1, re3, & ag2, re3, & ag3, re1, & ag3, re2, & \it Avg \\
Case order &  pa3 & pa3  & pa2 & pa1 & pa2 & pa1 & \it markedness\\
\midrule
NDA & 0.94 & 0.91 & 0.53 & 0.41 & 0.68 & 0.64 & \it0.69 \\
NAD & 0.64 & 0.55 & 0.84 & 0.71 & 0.60 & 0.62 & \it0.66 \\
DNA & 0.91 & 0.90 & 0.42 & 0.39 & 0.27 & 0.29 & \it0.53 \\
AND & 0.32 & 0.31 & 0.42 & 0.39 & 0.22 & 0.18 & \it0.31 \\
DAN & 0.53 & 0.55 & 0.28 & 0.21 & 0.76 & 0.72 & \it0.51 \\
ADN & 0.22 & 0.14 & 0.04 & 0.03 & 0.39 & 0.42 & \it0.21 \\
\it Avg plaus. & \it0.59 & \it0.56 & \it0.42 & \it0.36 & \it0.49 & \it0.48 & \it0.48 \\
\bottomrule
\end{tabular}
\caption{(1-2) accusative \mvss LSTM AUCs (higher is better) averaged by case order and ranked by increasing \textit{markedness} (NDA least marked) and decreasing \textit{plausibility} (ag1,re2,pa3 most plausible).}
\label{tab:lstm_1to2_acc}
\end{table*}

\begin{table*}[t]
\centering
\begin{tabular}{lccccccc}
\toprule
Role assignment / & ag1, re2, & ag2, re1& ag1, re3, & ag2, re3, & ag3, re1, & ag3, re2, & \it Avg \\
Case order &  pa3 & pa3  & pa2 & pa1 & pa2 & pa1 & \it markedness\\
\midrule
NDA & 0.99 & 0.99 & 0.49 & 0.44 & 0.76 & 0.75 & \it0.74 \\
NAD & 0.90 & 0.91 & 0.88 & 0.87 & 0.78 & 0.83 & \it0.86 \\
DNA & 0.99 & 0.99 & 0.56 & 0.46 & 0.72 & 0.68 & \it0.73 \\
AND & 0.91 & 0.92 & 0.80 & 0.83 & 0.59 & 0.63 & \it0.78 \\
DAN & 0.93 & 0.87 & 0.43 & 0.45 & 0.94 & 0.92 & \it0.76 \\
ADN & 0.98 & 0.97 & 0.46 & 0.48 & 0.96 & 0.94 & \it0.80 \\
\it Avg plaus. & \it0.95 & \it0.94 & \it0.60 & \it0.59 & \it0.79 & \it0.79 & \it0.78 \\
\bottomrule
\end{tabular}
\caption{(1-2) accusative \mvss BERT AUCs (higher is better) averaged by case order and ranked by increasing \textit{markedness} (NDA least marked) and decreasing \textit{plausibility} (ag1,re2,pa3 most plausible).}
\label{tab:bert_1to2_acc}
\end{table*}

\begin{table*}[t]
\centering
\begin{tabular}{lccccccc}
\toprule
Role assignment / & ag1, re2, & ag2, re1& ag1, re3, & ag2, re3, & ag3, re1, & ag3, re2, & \it Avg \\
Case order &  pa3 & pa3  & pa2 & pa1 & pa2 & pa1 & \it markedness\\
\midrule
NDA & 1.00 & 0.98 & 0.43 & 0.36 & 0.75 & 0.72 & \it0.71 \\
NAD & 0.92 & 0.86 & 0.92 & 0.93 & 0.74 & 0.82 & \it0.87 \\
DNA & 0.97 & 0.96 & 0.46 & 0.37 & 0.56 & 0.46 & \it0.63 \\
AND & 0.92 & 0.84 & 0.74 & 0.69 & 0.43 & 0.47 & \it0.68 \\
DAN & 0.51 & 0.46 & 0.24 & 0.30 & 0.79 & 0.82 & \it0.52 \\
ADN & 0.85 & 0.83 & 0.27 & 0.32 & 0.87 & 0.86 & \it0.67 \\
\it Avg plaus. & \it0.86 & \it0.82 & \it0.51 & \it0.49 & \it0.69 & \it0.69 & \it0.68 \\
\bottomrule
\end{tabular}
\caption{(1-2) accusative \mvss DistilBERT AUCs (higher is better) averaged by case order and ranked by increasing \textit{markedness} (NDA least marked) and decreasing \textit{plausibility} (ag1,re2,pa3 most plausible).}
\label{tab:distilbert_1to2_acc}
\end{table*}

\section{Models AUCs (1-2) dative \mvss}
\label{sec:1-2_dat_tables}
Tables \ref{tab:unigram_1to2_dat}, \ref{tab:bigram_1to2_dat}, \ref{tab:lstm_1to2_dat}, \ref{tab:bert_1to2_dat} and \ref{tab:distilbert_1to2_dat} show AUCs for (1-2) dative \mvs across all models. AUCs are averaged by case order and ranked by increasing \textit{markedness} (NDA least marked) and decreasing \textit{plausibility} (ag1,re2,pa3 most plausible) according to humans.

\begin{table*}[t]
\centering
\begin{tabular}{lccccccc}
\toprule
Role assignment / & ag1, re2, & ag2, re1& ag1, re3, & ag2, re3, & ag3, re1, & ag3, re2, & \it Avg \\
Case order &  pa3 & pa3  & pa2 & pa1 & pa2 & pa1 & \it markedness\\
\midrule
NDA & 0.99 & 0.99 & 0.64 & 0.64 & 0.60 & 0.64 & \it0.75 \\
NAD & 0.88 & 0.89 & 0.92 & 0.84 & 0.67 & 0.61 & \it0.80 \\
DNA & 0.83 & 0.85 & 0.54 & 0.58 & 0.61 & 0.50 & \it0.65 \\
AND & 0.68 & 0.80 & 0.67 & 0.69 & 0.46 & 0.49 & \it0.63 \\
DAN & 0.76 & 0.69 & 0.54 & 0.57 & 0.59 & 0.57 & \it0.62 \\
ADN & 0.68 & 0.66 & 0.55 & 0.51 & 0.57 & 0.60 & \it0.60 \\
\it Avg plaus. & \it0.80 & \it0.81 & \it0.64 & \it0.64 & \it0.58 & \it0.57 & \it0.67 \\
\bottomrule
\end{tabular}
\caption{(1-2) dative \mvss human AUCs (higher is better) averaged by case order and ranked by increasing \textit{markedness} (NDA least marked) and decreasing \textit{plausibility} (ag1,re2,pa3 most plausible).}
\label{tab:human_1to2_dat}
\end{table*}

\begin{table*}[t]
\centering
\begin{tabular}{lccccccc}
\toprule
Role assignment / & ag1, re2, & ag2, re1& ag1, re3, & ag2, re3, & ag3, re1, & ag3, re2, & \it Avg \\
Case order &  pa3 & pa3  & pa2 & pa1 & pa2 & pa1 & \it markedness\\
\midrule
NDA & 0.95 & 0.80 & 0.94 & 0.80 & 0.78 & 0.78 & \it0.84 \\
NAD & 0.80 & 0.67 & 0.93 & 0.81 & 0.75 & 0.78 & \it0.79 \\
DNA & 0.95 & 0.80 & 0.94 & 0.80 & 0.78 & 0.78 & \it0.84 \\
AND & 0.80 & 0.67 & 0.93 & 0.81 & 0.75 & 0.78 & \it0.79 \\
DAN & 0.81 & 0.77 & 0.97 & 0.93 & 0.95 & 0.97 & \it0.90 \\
ADN & 0.81 & 0.77 & 0.97 & 0.93 & 0.95 & 0.97 & \it0.90 \\
\it Avg plaus. & \it0.85 & \it0.75 & \it0.95 & \it0.85 & \it0.83 & \it0.84 & \it0.84 \\
\bottomrule
\end{tabular}
\caption{(1-2) dative \mvss unigram AUCs (higher is better) averaged by case order and ranked by increasing \textit{markedness} (NDA least marked) and decreasing \textit{plausibility} (ag1,re2,pa3 most plausible).}
\label{tab:unigram_1to2_dat}
\end{table*}

\begin{table*}[t]
\centering
\begin{tabular}{lccccccc}
\toprule
Role assignment / & ag1, re2, & ag2, re1& ag1, re3, & ag2, re3, & ag3, re1, & ag3, re2, & \it Avg \\
Case order &  pa3 & pa3  & pa2 & pa1 & pa2 & pa1 & \it markedness\\
\midrule
NDA & 0.92 & 0.96 & 0.83 & 0.80 & 0.87 & 0.83 & \it0.87 \\
NAD & 0.85 & 0.88 & 0.92 & 0.88 & 0.84 & 0.81 & \it0.86 \\
DNA & 0.96 & 0.87 & 0.87 & 0.79 & 0.75 & 0.74 & \it0.83 \\
AND & 0.77 & 0.71 & 0.87 & 0.80 & 0.57 & 0.62 & \it0.72 \\
DAN & 0.97 & 0.95 & 0.95 & 0.92 & 0.92 & 0.94 & \it0.94 \\
ADN & 0.71 & 0.69 & 0.63 & 0.61 & 0.69 & 0.61 & \it0.66 \\
\it Avg plaus. & \it0.86 & \it0.84 & \it0.85 & \it0.80 & \it0.77 & \it0.76 & \it0.81 \\
\bottomrule
\end{tabular}
\caption{(1-2) dative \mvss bigram AUCs (higher is better) averaged by case order and ranked by increasing \textit{markedness} (NDA least marked) and decreasing \textit{plausibility} (ag1,re2,pa3 most plausible).}
\label{tab:bigram_1to2_dat}
\end{table*}

\begin{table*}[t]
\centering
\begin{tabular}{lccccccc}
\toprule
Role assignment / & ag1, re2, & ag2, re1& ag1, re3, & ag2, re3, & ag3, re1, & ag3, re2, & \it Avg \\
Case order &  pa3 & pa3  & pa2 & pa1 & pa2 & pa1 & \it markedness\\
\midrule
NDA & 0.99 & 0.99 & 0.88 & 0.81 & 0.97 & 0.97 & \it0.93 \\
NAD & 0.97 & 0.92 & 0.96 & 0.92 & 0.91 & 0.94 & \it0.94 \\
DNA & 0.97 & 0.99 & 0.95 & 0.92 & 0.92 & 0.90 & \it0.94 \\
AND & 0.83 & 0.80 & 0.70 & 0.68 & 0.65 & 0.67 & \it0.72 \\
DAN & 0.96 & 0.95 & 0.88 & 0.92 & 0.74 & 0.85 & \it0.88 \\
ADN & 0.83 & 0.84 & 0.49 & 0.55 & 0.58 & 0.58 & \it0.65 \\
\it Avg plaus. & \it0.92 & \it0.92 & \it0.81 & \it0.80 & \it0.79 & \it0.82 & \it0.84 \\
\bottomrule
\end{tabular}
\caption{(1-2) dative \mvss LSTM AUCs (higher is better) averaged by case order and ranked by increasing \textit{markedness} (NDA least marked) and decreasing \textit{plausibility} (ag1,re2,pa3 most plausible).}
\label{tab:lstm_1to2_dat}
\end{table*}

\begin{table*}[t]
\centering
\begin{tabular}{lccccccc}
\toprule
Role assignment / & ag1, re2, & ag2, re1& ag1, re3, & ag2, re3, & ag3, re1, & ag3, re2, & \it Avg \\
Case order &  pa3 & pa3  & pa2 & pa1 & pa2 & pa1 & \it markedness\\
\midrule
NDA & 0.99 & 0.99 & 0.81 & 0.81 & 0.95 & 0.96 & \it0.92 \\
NAD & 0.98 & 1.00 & 0.91 & 0.90 & 0.99 & 0.97 & \it0.96 \\
DNA & 0.98 & 0.98 & 0.90 & 0.85 & 0.94 & 0.92 & \it0.93 \\
AND & 0.99 & 0.99 & 0.74 & 0.74 & 0.84 & 0.86 & \it0.86 \\
DAN & 0.98 & 0.98 & 0.83 & 0.86 & 0.81 & 0.80 & \it0.88 \\
ADN & 1.00 & 1.00 & 0.95 & 0.92 & 0.90 & 0.91 & \it0.95 \\
\it Avg plaus. & \it0.99 & \it0.99 & \it0.86 & \it0.85 & \it0.90 & \it0.90 & \it0.91 \\
\bottomrule
\end{tabular}
\caption{(1-2) dative \mvss BERT AUCs (higher is better) averaged by case order and ranked by increasing \textit{markedness} (NDA least marked) and decreasing \textit{plausibility} (ag1,re2,pa3 most plausible).}
\label{tab:bert_1to2_dat}
\end{table*}

\begin{table*}[t]
\centering
\begin{tabular}{lccccccc}
\toprule
Role assignment / & ag1, re2, & ag2, re1& ag1, re3, & ag2, re3, & ag3, re1, & ag3, re2, & \it Avg \\
Case order &  pa3 & pa3  & pa2 & pa1 & pa2 & pa1 & \it markedness\\
\midrule
NDA & 0.98 & 0.97 & 0.80 & 0.81 & 0.93 & 0.94 & \it0.90 \\
NAD & 1.00 & 0.98 & 0.79 & 0.84 & 0.95 & 0.98 & \it0.92 \\
DNA & 0.97 & 0.94 & 0.90 & 0.86 & 0.89 & 0.86 & \it0.90 \\
AND & 0.99 & 0.95 & 0.71 & 0.68 & 0.77 & 0.83 & \it0.82 \\
DAN & 0.90 & 0.84 & 0.63 & 0.73 & 0.56 & 0.70 & \it0.73 \\
ADN & 0.97 & 0.96 & 0.68 & 0.78 & 0.78 & 0.78 & \it0.83 \\
\it Avg plaus. & \it0.97 & \it0.94 & \it0.75 & \it0.78 & \it0.81 & \it0.85 & \it0.85 \\
\bottomrule
\end{tabular}
\caption{(1-2) dative \mvss DistilBERT AUCs (higher is better) averaged by case order and ranked by increasing \textit{markedness} (NDA least marked) and decreasing \textit{plausibility} (ag1,re2,pa3 most plausible).}
\label{tab:distilbert_1to2_dat}
\end{table*}

\end{document}